\documentclass[conference]{IEEEtran}
\usepackage{cite}
\usepackage{amsmath,amssymb,amsfonts}
\usepackage{algorithmic}
\usepackage{graphicx}
\usepackage{textcomp}
\usepackage{hyperref}
\usepackage{xcolor}
\usepackage{caption}
\def\BibTeX{{\rm B\kern-.05em{\sc i\kern-.025em b}\kern-.08em
    T\kern-.1667em\lower.7ex\hbox{E}\kern-.125emX}}
\begin{document}

\title{Data Augmentation for Automated Essay Scoring using Transformer Models \\}

\author{\IEEEauthorblockN{Kshitij Gupta}
\IEEEauthorblockA{\textit{Department of Electrical and Electronics Engineering} \\
\textit{BITS Pilani, Pilani Campus}\\
Pilani, India \\
mailguptakshitij@gmail.com}
}

\maketitle
\begin{abstract}
    Automated essay scoring is one of the most important problem in Natural Language Processing.  It has been explored for a number of years, and it remains partially solved. In addition to its economic and educational usefulness, it presents research problems. Transfer learning has proved to be beneficial in NLP. Data augmentation techniques have also helped build state-of-the-art models for automated essay scoring. Many works in the past have attempted to solve this problem by using RNNs, LSTMs, etc. This work examines the transformer models like BERT, RoBERTa, etc. We empirically demonstrate the effectiveness of transformer models and data augmentation for automated essay grading across many topics using a single model.
\end{abstract}

\section{Introduction}
As a result of the COVID-19 pandemic, online schooling system became necessary. From elementary schools to colleges, almost all educational institutions have adopted the online education system. The majority of automated evaluations are accessible for multiple-choice questions, but evaluating short and essay type responses remains unsolved since, unlike multiple-choice questions, there is no one correct solution for these kind of questions. It is an essential education-related application that employs NLP and machine learning methodologies. It is difficult to evaluate essays using basic computer languages and methods such as pattern matching and language processing.

Among the most important pedagogical uses of NLP is automated essay scoring (AES), the technique of using a system to score short and essay type questions without manual assistance. Initiated by Page's [1966] groundbreaking work on the Project Essay Grader system, this area of study has seen continuous activity ever since. The bulk of AES research has been on holistic scoring, which provides a quantitative summary of an essay's quality in a single number. At least two factors contribute to this concentration of effort. To begin with, learning-based holistic scoring systems may make use of publically accessible corpora that have been manually annotated with holistic scores. Second, there is a market for holistic scoring algorithms because they may streamline the arduous process of manually evaluating the millions of essays for tests like GRE, IELTS, SAT.

Past research on automated essay grading has included training models for essays for which training data is available and those models are topic specific. This model is trained on all the topics thus could be used for assessment of essays of all those topics without training model specific for each topic. This would be useful in the scenario where we did not have enough data to train a model that is specific to a particular topic, but we still needed to evaluate essays on that topic. Therefore, in order to assess them, We may utilize a model that has been trained on essays on a variety of topics and a tiny amount of data on the topic for which we need to develop a model, which will then be fine-tuned using the limited data available on the subject being assessed.

This paper is organized as follows: In Section II, we explore pertinent prior research on automated essay scoring; in Section III, we cover experimental setup; and in Section IV, we describe our methodology for augmenting essay data. In Section V, we give the results and analysis of the automated essay grading model. Section VI comprises of conclusion and future work for Automated Essay Scoring.

\section{Related Works}
  Project Essay Grader (PEG) by \cite{b1} started the research on Automated essay scoring. Shermis (2001)\cite{b2} improved the PEG system by incorporating the grammatical features as well in the evaluation. Around the turn of century, great majority of essay scoring systems used conventional methods like latent semantic analysis by Foltz (1999)\cite{b3}, as pattern matching and statistical analysis like Bayesian Essay Test Scoring System by \cite{b4}. These systems employ natural language processing (NLP) approaches that concentrate on grammar, content to determine an essay's score.

Multiple studies studied AES systems, from the earliest to the most recent. Blood (2011)\cite{b6} reviewed the PEG literature from 1984 to 2010, it has discussed just broad features of AES systems, such as ethical considerations and system performance. However, they have not addressed the implementation aspect, nor has a comparison research been conducted, nor have the real problems of AES systems been highlighted. 

After 2014,  Automated grading systems like as those by \cite{b5} and others, employed deep learning approaches to induce syntactic and semantic characteristics, producing greater outcomes than previous systems. Burrows (2015)\cite{b7} reviewed on aspects, including datasets, NLP approaches, model construction, model grading, model assessment, and model efficacy. Ke (2019)\cite{b8}, Hussein (2019)\cite{b9} and Klebanov (2020)\cite{b10} offered overview of the AES system.

Ramesh (2019)\cite{b19} offered us a comprehensive summary of all accessible datasets and the machine learning algorithms used to grade essays. Recently Park (2022)\cite{b23} tried using GAN's for evaluation of essays. Our motivation of using Deep-learning models for automated essay scoring was due to recent study by Elijah (2020)\cite{b20} and Wang (2022)\cite{b25}, it showed us that using BERT based models it certainly improves the accuracy of the Automated essay scoring. 

Jong (2022)\cite{b24} demonstrated the effectiveness of data augmentation techniques on automated essay scoring. Ludwig (2021)\cite{b21}, Ormerod (2021)\cite{b22} and Sethi (2022)\cite{b26} examined the transformer based models on automated essay scoring, these recent findings prompted us to test our data augmentation strategy on transformer-based models for automated essay grading.

\section{Experimental Setup}
For the purpose of this study, we will constrain the study and experiments to ASPA1 dataset. The \href{https://www.kaggle.com/c/asap-aes}{Automated Student Assessment Prize(ASAP1)}\footnote{https://www.kaggle.com/c/asap-aes} corpus was released as part of a Kaggle competition in 2012. This corpus is used for holistic scoring of essays and consists of essays from eight topics with around 3000 essays on each of the topic. Each topic had a different scoring method, so, we normalized of each essay score from 0 to 10 so that we could train all the data together.

We run our experiments using the BERT, RoBERTa, ALBERT, DistilBERT, XLM-RoBERTa,  implementation available in the Simple Transformers library provided by \href{https://simpletransformers.ai/}{Thilina Rajapakse}\footnote{https://simpletransformers.ai/}. We run our experiment on Google Colab. For baseline trials, we consider the essays which are not supplemented with our data augmentation technique and trained on models based on Transformers. 

During training, we additionally fine-tuned our parameters by adjusting the parameters like learning rate, weight decay rate, etc. For the purpose of evaluating our model, we will use the accuracy score measure given by the \href{https://scikit-learn.org/stable/index.html}{Scikit library}\footnote{https://scikit-learn.org/stable/index.html}. Since we are approaching our issue as a multi-label classification, the accuracy metric is often used to evaluate multi-label classifications. 
\begin{figure}[]
      \includegraphics[scale=0.2]{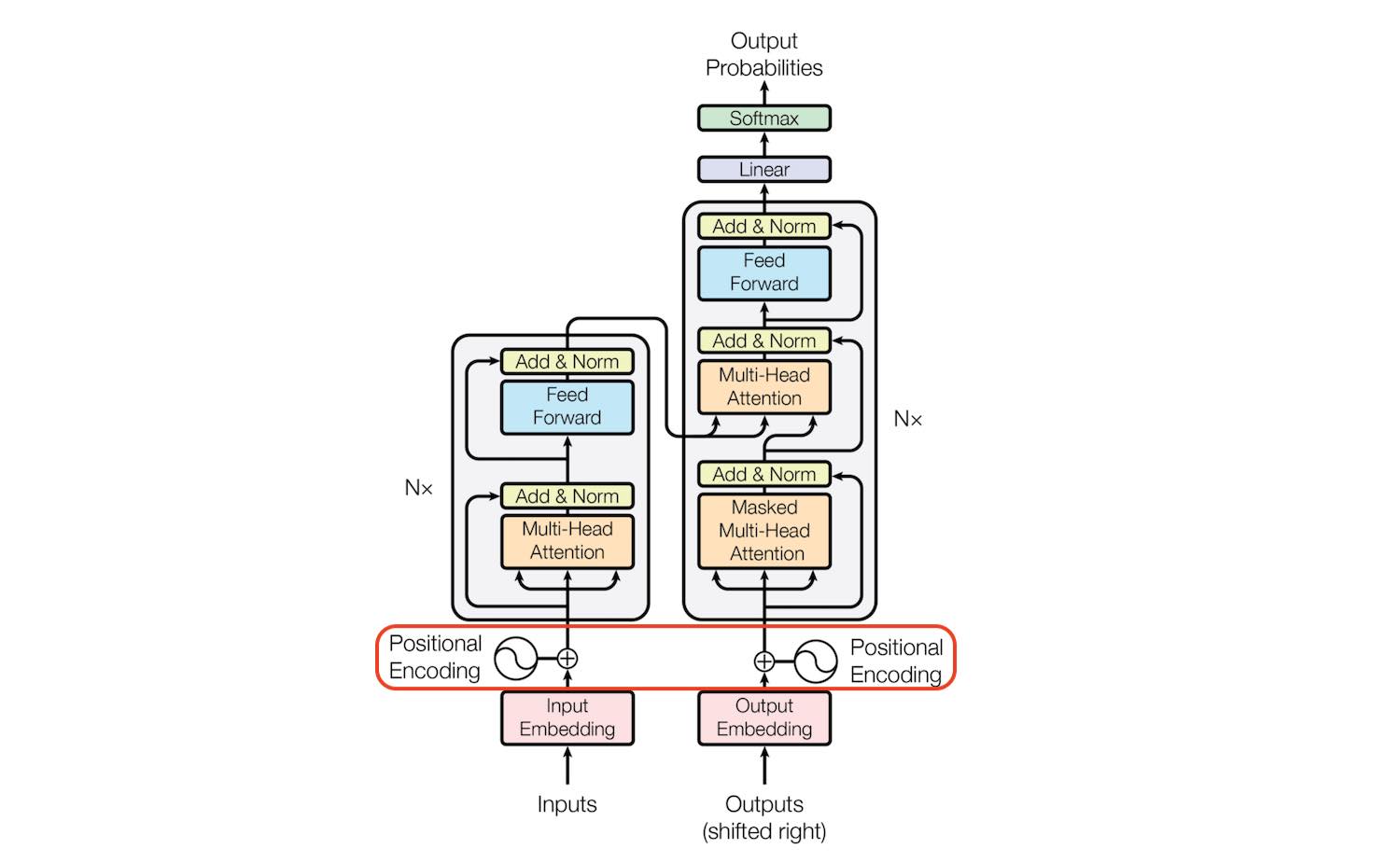}
      \caption{Transformers Architecture}
      \label{fig:transformers}
  \end{figure}

\section{Methodology}
\subsection{Large Pre-trained models}
The models which we used for training are based on the Transformers(Fig. \ref{fig:transformers}) architecture introduced by \cite{b11}. The Transformer's architecture follows an encoder-decoder structure. 

Given below is the brief description of each of the model which we are using for training:
\subsubsection{\text{BERT}} 
The Bidirectional Encoder Representations (BERT) introduced by \cite{b12} is a deep learning model in which every output element is linked to every input element and the weightings between them are dynamically determined depending on their relationship. BERT is pre-trained on two tasks: Masked Language Modeling and Next Sentence Prediction.

\subsubsection{\text{RoBERTa}}
Robustly optimized BERT Pre-training Approach (RoBERTa) introduced by \cite{b13} builds upon BERT's language masking method, in which the system learns to anticipate purposely masked bits of text inside unannotated language samples. RoBERTa changes critical hyperparameters in BERT, such as eliminating BERT's next-sentence pretraining target and training with much bigger mini-batches and learning rates. This enables RoBERTa to outperform BERT at the masked language modeling goal and improves the performance of subsequent tasks.
\subsubsection{\text{ALBERT}}
ALBERT, introduced by \cite{b14} is an encoder-decoder based model with self-attention at the encoder and attention to encoder outputs at the decoder end. It is a modified version of BERT and it stands for "A Lite BERT". It builds on parameter sharing, embedding factorization, and sentence order prediction(SOP).
\subsubsection{\text{DistilBERT}}
DistilBERT, introduced by \cite{b15} aims to optimize the training by reducing the size of BERT and increasing the speed of BERT—all while trying to retain as much performance as possible. Specifically, DistilBERT is smaller than the original BERT-base model, is faster than it, and retains its functionality.
\subsubsection{\text{XLM-RoBERTa}}
XLM-RoBERTa, Unsupervised Cross-lingual Representation Learning at Scale, introduced by \cite{b16} is a scaled cross-lingual sentence encoder. It is trained on 2.5 TB of data across 100 languages filtered from Common Crawl. XLM-RoBERTa achieves state-of-the-art results on multiple cross-lingual benchmarks.
\begin{figure*}[t!]
      \centering
      \includegraphics[scale=0.55]{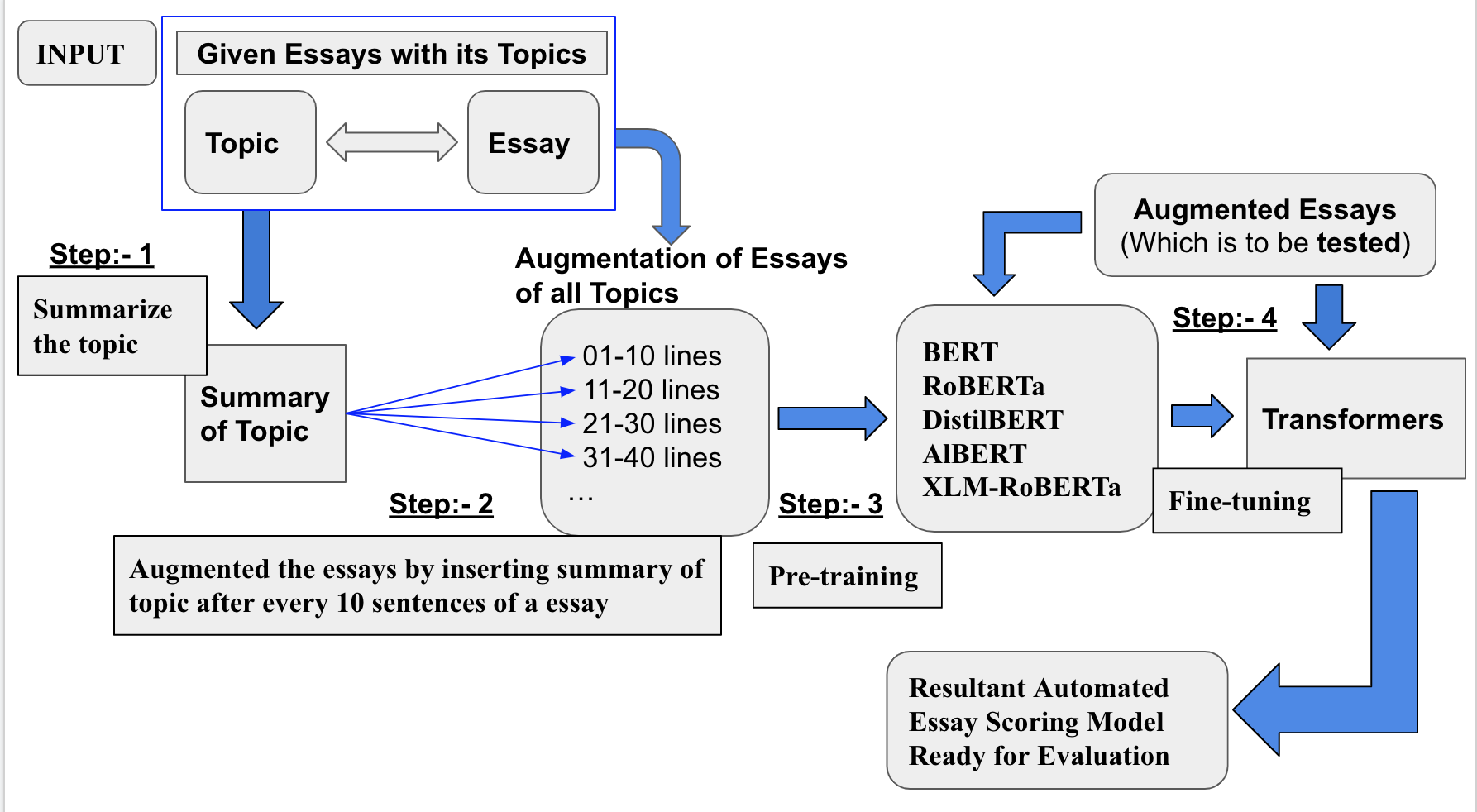}
      \caption{Augmentation Technique}
      \label{fig:augmentation}
  \end{figure*}
\subsection{Data Augmentation}
Researchers have attempted to use several RNNs and LSTMs as training models for automated essay scoring. However, the fundamental disadvantage of such models is that they are topic-specific, and we want to construct an automated essay scoring system that can perform well not just on subjects for which we have an abundance of data but also on subjects for which we have a limited amount of data. Now, we want to augment the essay so that it can accurately assess a essay on a different topic for which we have a very small amount of data.

When training a model, we add each essay with its topic at certain intervals. We are including essay topics after an interval because essays are lengthy, and if we train only on essays, the trained model will be topic-specific. In order to construct a more robust model, we append the essay's subject to each essay so that the training model may learn the relationship between the essay's topic and the essay itself. This is so that it accurately grades essays on a different topic, which we are using majorly for fine-tuning since we have very little data available to us and we can not train topic-specific models using this less amount of data.

We cannot add a complete topic to an essay because sometimes essay topics are quite large. Hence, we summarize the topics of essays using the summarization pipeline provided by \cite{b17} implementation of BART, which was introduced by \cite{b18}. Now, the second issue is, after how many lines should we put the subject for optimum precision? We conducted extensive data trials by inserting them at different places.

After a comprehensive investigation of several transformer-based models and essay subject insertions, we determined that the tenth place is optimal for inserting the topic. It implies that after every tenth line, a summary of the subject is added to the essay. It may be due to the fact that the model struggles with lengthy text classification; thus, we keep this in mind by inserting topics at regular intervals. Fig. \ref{fig:augmentation} shows the complete data augmentation of essay. Now that the topic has been inserted into the essay, the data is modified for training transformer models. The next section provides a study of modified essays.

\begin{table*}[t]
\caption{\textbf{Sklearn \href{https://scikit-learn.org/stable/modules/generated/sklearn.metrics.accuracy_score.html?highlight=accuracy_score}{Accuracy Score} of Mutli-Label Classification of all the four topics in the testing set}}
\label{tab:table1}
\begin{minipage} {.5\linewidth}
\centering
    \captionsetup{labelformat=empty}
    \caption{Topic 1}
    \begin{tabular}{ |p{2cm}||p{2cm}|p{2cm}| } 
 \hline
  \textbf{Transformer Models}& \textbf{Unaugmented Data}& \textbf{Augmented Data} \\
 \hline
 \hline
  LSTM & 30.8\% & 38.3\% \\
  \hline
  XLM-RoBERTa & 47.4\% & 59.2\% \\ 
  \hline
  \textbf{RoBERTa} & \textbf{51.1}\% &  \textbf{60.4}\% \\ 
 \hline
  ALBERT & 48.7\% & 59.7\% \\ 
  \hline 
  DistilBERT & 50.3\% & 58.1\% \\
  \hline
  \textbf{BERT} & \textbf{50.7}\% & \textbf{60.2}\% \\ 
 \hline
\end{tabular}
\end{minipage}\hfill
\begin{minipage}{.5\linewidth}
\centering
\captionsetup{labelformat=empty}
\caption{Topic 2}
\begin{tabular}{ |p{2cm}||p{2cm}|p{2cm}| } 
 \hline
  \textbf{Transformer Models}& \textbf{Unaugmented Data}& \textbf{Augmented Data} \\
 \hline
 \hline
  LSTM & 29.4\% & 37.6\% \\
  \hline
  XLM-RoBERTa & 47.8\% & 58.6\% \\ 
  \hline
  \textbf{RoBERTa} & \textbf{50.9}\% &  \textbf{60.8}\% \\ 
 \hline
  ALBERT & 49.6\% & 60.2\% \\ 
  \hline 
  DistilBERT & 49.8\% & 58.8\% \\
  \hline
  \textbf{BERT} & \textbf{50.2}\% & \textbf{61.3}\% \\ 
 \hline
\end{tabular}
\end{minipage}
\\
\begin{minipage}{.5\linewidth}
\centering
\captionsetup{labelformat=empty}
\caption{Topic 3}
\begin{tabular}{ |p{2cm}||p{2cm}|p{2cm}| } 
 \hline
  \textbf{Transformer Models}& \textbf{Unaugmented Data}& \textbf{Augmented Data} \\
 \hline
 \hline
  LSTM & 30.3\% & 38.1\% \\
  \hline
  XLM-RoBERTa & 46.9\% & 58.9\% \\ 
  \hline
  \textbf{RoBERTa} & \textbf{51.3}\% &  \textbf{61.3}\% \\ 
 \hline
  ALBERT & 49.2\% & 59.9\% \\ 
  \hline 
  DistilBERT & 49.4\% & 58.9\% \\
  \hline
  \textbf{BERT} & \textbf{49.3}\% & \textbf{61.5}\% \\ 
 \hline
\end{tabular}
\end{minipage}\hfill
\begin{minipage}{.5\linewidth}
\centering
\captionsetup{labelformat=empty}
\caption{Topic 4}
\begin{tabular}{ |p{2cm}||p{2cm}|p{2cm}| } 
 \hline
  \textbf{Transformer Models}& \textbf{Unaugmented Data}& \textbf{Augmented Data} \\
 \hline
 \hline
  LSTM & 30.3\% & 38.9\% \\
  \hline
  XLM-RoBERTa & 46.9\% & 58.4\% \\ 
  \hline
  \textbf{RoBERTa} & \textbf{50.5}\% &  \textbf{61.4}\% \\ 
 \hline
  ALBERT & 47.6\% & 59.3\% \\ 
  \hline 
  DistilBERT & 51.1\% & 58.2\% \\
  \hline
  \textbf{BERT}& \textbf{50.3}\% & \textbf{61.7\%} \\ 
 \hline
\end{tabular}
\end{minipage}

\end{table*}

\section{Results and Analysis}
The ASAP1 dataset contains around 17K essays on eight topics. We are using that data for pre-training by augmenting those essays using our technique. For fine tuning 
For research and testing purposes, we used this \href{https://drive.google.com/drive/folders/1lphQ7IS30vNZtIdOCe5eCoCAWbbCblEx?usp=sharing}{dataset}. This dataset consists of 1241 essays on four subjects. We fine-tuned and tested our models on each subject individually. We used around two-thirds of the above mentioned dataset for training and fine-tuning, and the remaining one-third for testing.

We followed a very simple yet state-of-the-art modeling technique for multi-label classification using transformer models. We bucketed scores into each interval class, resulting in 11 buckets. These 11 classes correspond to a score from 0 to 10. Our methodology assigns each essay to a particular category. If an essay is categorized by my model as being in \textcolor{cyan}{Bucket 6}, then it receives a score of \textcolor{cyan}{5}. Models like BERT, RoBERTa, ALBERT, DistilBERT, and XLM-RoBERT\footnote{we used base models for our experiment} were used to teach augmented essays how to recognize multiple labels.

We approached this issue in the same manner as sentiment analysis, which utilizes classification algorithms and yields extremely positive results. We have discovered that using this data augmentation training method contributes to an improvement in accuracy as referenced in Table \ref{tab:table1}. From these results, it is quite evident that BERT and RoBERTa outperform other models, although by a small margin. The analysis of these findings demonstrates that utilizing Transformer-based models is considerably superior than using LSTMs.

When used on top of huge pre-trained classification models, our data augmentation strategies significantly enhance the performance of automated essay grading. The accuracy of all those pre-trained models increases after applying our augmentation technique. We believe this performance is because after mixing a summary of the topic with each essay, it encourages the internal representation of each essay to align with the topic, so that when we test it on a essay with a different topic after fine tuning, it checks for the alignment between the topic and the essay as a result of training and fine-tuning and grades it accordingly. 

\section{Conclusion and Future work}
In this paper, we automated essay grading using transformer based models and an augmentation approach. The conclusion are as follows:
\begin{itemize}
    \item Pre-trained transformer-based models BERT, RoBERTa, ALBERT, DistilBERT, and XLM-RoBERTa are very proficient at Automated Essay Scoring.
    \item Data augmentation approaches might further enhance its performance for analyzing lengthier resources such as essays to attain accurate essay scores.
\end{itemize}
In the future, we'll come up with a better way to include elements that are relevant to the topic instead of just a summary, so that training with this data will lead to a more accurate model for automatically grading essays.


\begin{thebibliography}{00}
\bibitem{b1} Ajay HB, Tillett PI, Page EB (1973) Analysis of essays by computer (AEC-II) (No. 8-0102). Washington, DC: U.S. Department of Health, Education, and Welfare, Office of Education, National Center for Educational Research and Development.
\bibitem{b2} Shermis MD, Mzumara HR, Olson J, Harrington S (2001) On-line grading of student essays: PEG goes on the World Wide Web. Assess Eval High Educ 26(3):247–259.
\bibitem{b3} Foltz PW, Laham D, Landauer TK (1999) The Intelligent Essay Assessor: Applications to Educational Technology. Interactive Multimedia Electronic Journal of Computer-Enhanced Learning, 1, 2, http://imej.wfu.edu/articles/1999/2/04/index.asp.
\bibitem{b4} Rudner, L. M.,  Liang, T. (2002). Automated essay scoring using Bayes' theorem. The Journal of Technology, Learning and Assessment, 1(2).
\bibitem{b5} Dong F, Zhang Y, Yang J (2017) Attention-based recurrent convolutional neural network for automatic essay scoring. In: Proceedings of the 21st Conference on Computational Natural Language Learning (CoNLL 2017) p 153–162.
\bibitem{b6} Blood, I. (2011). Automated essay scoring: a literature review. Studies in Applied Linguistics and TESOL, 11(2).
\bibitem{b7} Burrows S, Gurevych I, Stein B (2015) The eras and trends of automatic short answer grading. Int J Artif Intell Educ 25:60–117. https://doi.org/10.1007/s40593-014-0026-8.
\bibitem{b8} Ke Z, Ng V (2019) “Automated essay scoring: a survey of the state of the art.” IJCAI
\bibitem{b9} Hussein, M. A., Hassan, H.,  Nassef, M. (2019). Automated language essay scoring systems: A literature review. PeerJ Computer Science, 5, e208.
\bibitem{b10} Klebanov, B. B., Madnani, N. (2020, July). Automated evaluation of writing–50 years and counting. In Proceedings of the 58th Annual Meeting of the Association for Computational Linguistics (pp. 7796–7810).
\bibitem{b11} Vaswani, Ashish, Noam Shazeer, Niki Parmar, Jakob Uszkoreit, Llion Jones, Aidan N. Gomez, Łukasz Kaiser, and Illia Polosukhin. "Attention is all you need." Advances in neural information processing systems 30 (2017).
\bibitem{b12} Jacob Devlin, Ming-Wei Chang, Kenton Lee, and Kristina Toutanova. 2019. BERT: Pre-training of Deep Bidirectional Transformers for Language Understanding. In Proceedings of the 2019 Conference of the North American Chapter of the Association for Computational Linguistics: Human Language Technologies, Volume 1 (Long and Short Papers), pages 4171–4186, Minneapolis, Minnesota. Association for Computational Linguistics.
\bibitem{b13} Liu, Yinhan, Ott, Myle, Goyal, Naman, Du, Jingfei, Joshi, Mandar, Chen, Danqi, Levy, Omer, Lewis, Mike, Zettlemoyer, Luke and Stoyanov, Veselin RoBERTa: A Robustly Optimized BERT Pretraining Approach. (2019). , cite arxiv:1907.11692.
\bibitem{b14} Lan, Zhenzhong, Chen, Mingda, Goodman, Sebastian, Gimpel, Kevin, Sharma, Piyush and Soricut, Radu. "ALBERT: A Lite BERT for Self-supervised Learning of Language Representations.." Paper presented at the meeting of the ICLR, 2020.
\bibitem{b15} Sanh, Victor, Lysandre Debut, Julien Chaumond, and Thomas Wolf. "DistilBERT, a distilled version of BERT: smaller, faster, cheaper and lighter." arXiv preprint arXiv:1910.01108 (2019).
\bibitem{b16} Conneau, Alexis, Kartikay Khandelwal, Naman Goyal, Vishrav Chaudhary, Guillaume Wenzek, Francisco Guzmán, Edouard Grave, Myle Ott, Luke Zettlemoyer, and Veselin Stoyanov. "Unsupervised cross-lingual representation learning at scale." arXiv preprint arXiv:1911.02116 (2019).
\bibitem{b17} Wolf, Thomas, Lysandre Debut, Victor Sanh, Julien Chaumond, Clement Delangue, Anthony Moi, Pierric Cistac et al. "Huggingface's transformers: State-of-the-art natural language processing." arXiv preprint arXiv:1910.03771 (2019)
\bibitem{b18} Lewis, Mike, Yinhan Liu, Naman Goyal, Marjan Ghazvininejad, Abdelrahman Mohamed, Omer Levy, Ves Stoyanov, and Luke Zettlemoyer. "Bart: Denoising sequence-to-sequence pre-training for natural language generation, translation, and comprehension." arXiv preprint arXiv:1910.13461 (2019).
\bibitem{b19} Ramesh, Dadi, and Suresh Kumar Sanampudi. "An automated essay scoring systems: a systematic literature review." Artificial Intelligence Review (2021): 1-33.
\bibitem{b20} Elijah Mayfield and Alan W Black. 2020. Should You Fine-Tune BERT for Automated Essay Scoring?. In Proceedings of the Fifteenth Workshop on Innovative Use of NLP for Building Educational Applications, pages 151–162, Seattle, WA, USA → Online. Association for Computational Linguistics.
\bibitem{b21} Ludwig, Sabrina, et al. "Automated essay scoring using transformer models." Psych 3.4 (2021): 897-915.
\bibitem{b22} Ormerod, Christopher M., Akanksha Malhotra, and Amir Jafari. "Automated essay scoring using efficient transformer-based language models." arXiv preprint arXiv:2102.13136 (2021).
\bibitem{b23} Park, Yo-Han, et al. "EssayGAN: Essay Data Augmentation Based on Generative Adversarial Networks for Automated Essay Scoring." Applied Sciences 12.12 (2022): 5803.
\bibitem{b24} Jong, You-Jin, Yong-Jin Kim, and Ok-Chol Ri. "Improving Performance of Automated Essay Scoring by using back-translation essays and adjusted scores." Mathematical Problems in Engineering 2022 (2022).
\bibitem{b25} Wang, Yongjie, et al. "On the Use of BERT for Automated Essay Scoring: Joint Learning of Multi-Scale Essay Representation." arXiv preprint arXiv:2205.03835 (2022).
\bibitem{b26} Sethi, Angad, and Kavinder Singh. "Natural Language Processing based Automated Essay Scoring with Parameter-Efficient Transformer Approach." 2022 6th International Conference on Computing Methodologies and Communication (ICCMC). IEEE, 2022.

\end{thebibliography}
\end{document}